\documentclass[lettersize,journal]{IEEEtran}
\usepackage{amsmath,amsfonts}
\usepackage{algorithmic}
\usepackage{algorithm}
\usepackage{array}
\usepackage[caption=false,font=normalsize,labelfont=sf,textfont=sf]{subfig}
\usepackage{textcomp}
\usepackage{stfloats}
\usepackage{url}
\usepackage{verbatim}
\usepackage{multirow}
\usepackage{booktabs}
\usepackage{graphicx}
\usepackage{amssymb}
\usepackage{caption}   
\usepackage{cite}
\usepackage[hidelinks]{hyperref}

\begin{document}

\title{\textbf{ChicGrasp}: Imitation‑Learning based Customized Dual‑Jaw Gripper Control for Delicate, Irregular Bio-products Manipulation}

\author{Amirreza Davar, Zhengtong Xu, Siavash Mahmoudi, Pouya Sohrabipour, Chaitanya Pallerla, Yu She, Wan Shou, Philip Crandall, Dongyi Wang

\thanks{ A. Davar, S. Mahmoudi, P. Sohrabipour, C. Pallerla, W. Shou, P. Crandall and D. Wang are with the University of Arkansas, Fayetteville, AR, USA, 72701, Z. Xu, Y. She are with Purdue University, West Lafayette, IN, USA, 47907}}



\markboth{Journal of \LaTeX\ Class Files,~Vol.~XX, No.~X}%
{Shell \MakeLowercase{\textit{et al.}}: A Sample Article Using IEEEtran.cls for IEEE Journals}


\maketitle


\begin{abstract}
Automated poultry processing lines still rely on humans to lift slippery, easily bruised carcasses onto a shackle conveyor. Deformability, anatomical variance, and strict hygiene rules make conventional suction and scripted motions unreliable. We present \emph{ChicGrasp}, an end-to-end hardware–software co-design for this task. An independently actuated dual-jaw pneumatic gripper clamps both chicken legs, while a conditional diffusion-policy controller, trained from only 50 multi-view teleoperation demonstrations (RGB + proprioception), plans 5-DoF end-effector motion, which includes jaw commands in one shot. On individually presented raw broiler carcasses, our system achieves a 40.6\% grasp-and-lift success rate and completes the pick-to-shackle cycle in 38 s, whereas state-of-the-art implicit behaviour cloning (IBC) and LSTM-GMM baselines fail entirely. All CAD, code, and datasets will be open-source. ChicGrasp shows that imitation learning can bridge the gap between rigid hardware and variable bio-products, offering a reproducible benchmark and a public dataset for researchers in agricultural engineering and robot learning.

\end{abstract}



\section{Introduction}



\begin{figure}[!h]
    \centering
        \includegraphics[width=3in]{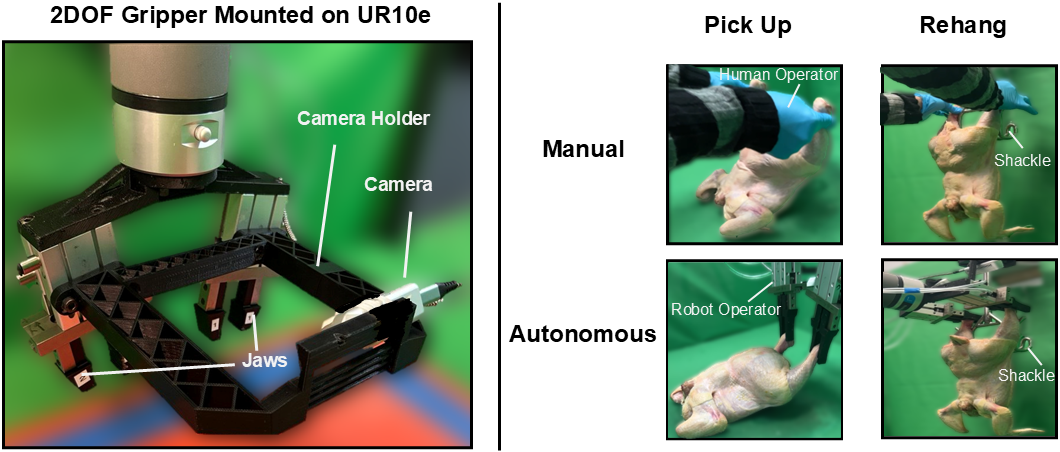}
        \caption{(Left) The robotic gripper equipped with two jaws and wrist camera designed for autonomous poultry handling. (Right) Comparison between the current industry-standard manual process and our autonomous approach using imitation learning to pick up and rehang poultry.}
    \label{Fig1_setup}
\end{figure}

\IEEEPARstart{R}{obots} and intelligent agents are increasingly deployed in unstructured, dynamic environments where manual programming struggles to capture the intricacies of real-world tasks \cite{osa2018algorithmic}. Poultry meat is the predominant protein for U.S. consumers, with broiler chicken production expected to exceed 21 million metric tons in 2025\cite{usda2024livestock}. However, some procedures in the existing poultry processing operation, such as hanging chilled carcasses onto moving shackle lines, remain heavily reliant on human labor, which makes this industry prone to risks of labor shortages and product contamination \cite{hafez1999poultry, rouger2017bacterial}. Figure \ref{Fig1_setup} illustrates our proposed robotic system in comparison to the current manual poultry hanging process.

\begin{figure*}[!h]
    \centering
        \includegraphics[width=5in]{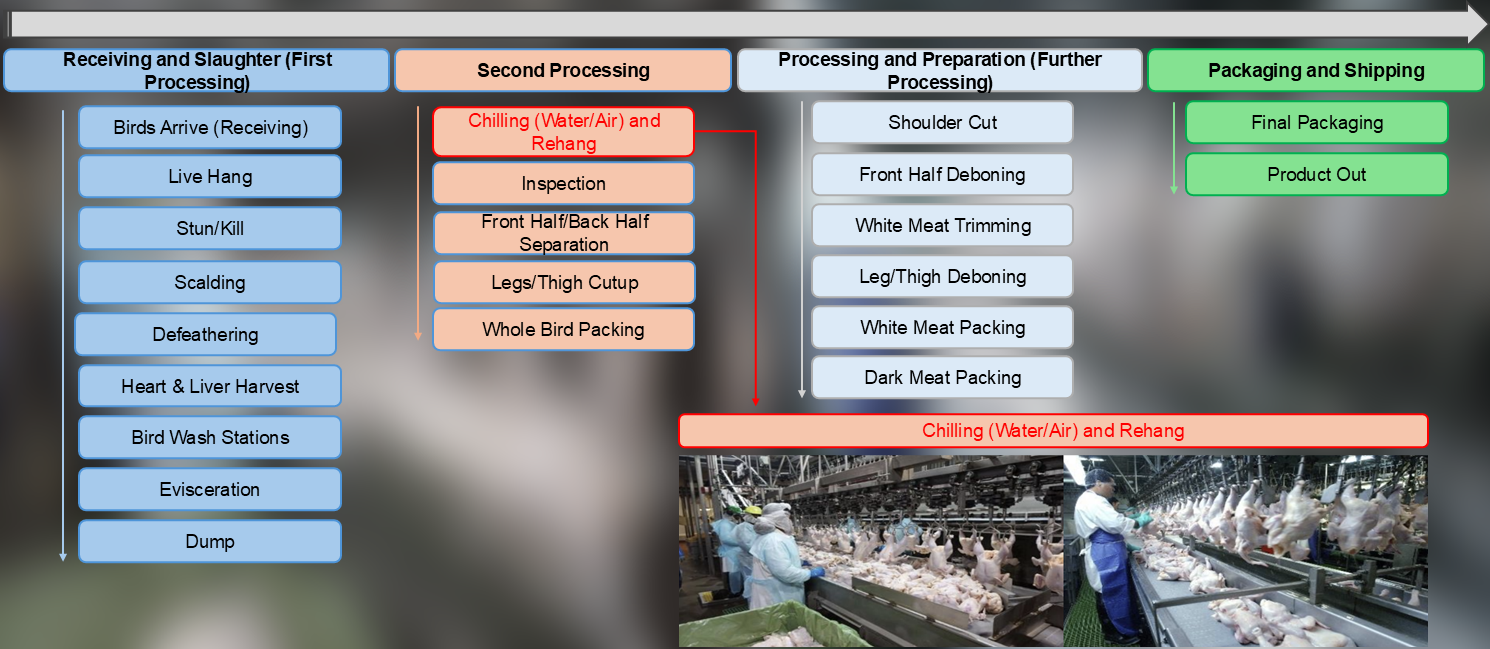}
        \caption{Poultry processing workflow from receiving to packaging which highlights the targeted automation in the chilling and rehang stage.}
    \label{Poultry processing workflow}
\end{figure*}

The operating lines of a typical poultry processing facility can be divided into three key stages: First Processing, Second Processing, and Further Processing, as depicted in Figure \ref{Poultry processing workflow}. First Processing includes stunning, de-feathering, and evisceration, while Second Processing focuses on cutting and separating different poultry parts. Finally, the Further Processing stage involves advanced operations such as deboning, trimming, and preparing poultry products for final retail packaging and shipping \cite{barbut2022quality, OSHA_PoultryProcessing}.


Early research on chicken processing and meat handling was limited to archaeological and anatomical studies, focusing on bone analysis rather than automation or robotic manipulation. Potts \cite{potts1985barebones} provides one such early investigation, emphasizing skeletal examination over automated handling. However, as the demand for improved efficiency in poultry processing grew, research shifted to adopting advanced sensing and automation techniques to improve the processing efficiency. 

Recently, researchers have explored AI methods focused on sensing and decision-making in the meat processing industry. Tran et al. \cite{tran2024carcassformer} developed a deep learning-based classification framework (CarcassFormer) for non-invasive quality inspection, but their work did not address the physical challenges of handling soft foods like poultry products. Nyalala et al. \cite{nyalala2024volume} investigated volume estimation techniques using depth imaging and machine learning for improved grading accuracy of chicken carcasses. Vision-guided robotic systems for automating poultry processing have also been evaluated. Xu et al. \cite{xu2023robotization} emphasized the role of robotic vision perception and cutting technologies in the meat industry, yet much of their work focused on the initial sensing rather than the adaptive manipulation portion. GRIBBOT system introduced by Misimi et al. \cite{misimi2016gribbot}, utilized a 3D imaging and a compliant gripper to automate chicken fillet harvesting. However, the GRIBBOT gripper could only follow predefined trajectories and did not incorporate learning-based methods to manage variations in easily deformable products. Joffe et al. \cite{joffe2019pose} explored robotic picking for poultry processing, using a deep learning-based pose estimation to grasp and reorient whole chicken carcasses randomly piled in a bin with a suction cup gripper. Although Joffe's deep learning approach improved subsequent automation steps, it still lacked an adaptive framework to continuously refine the grasping strategies. Furthermore, suction cup-type grippers are not reliable when grasping irregular, wet, or easily deformable surfaces, as they cause visible damage to the product surface, affecting its appearance \cite{pettersson2010design}. The non-uniform, slippery skin of poultry may prevent suction cup grippers from achieving a sufficient seal to successfully lift a whole chicken carcass. Suction cup grippers are inherently unsuitable for grasping specific anatomical features like legs, which vary in position and orientation from one bird to another. As a result, this method lacks the precision required for leg-based rehang tasks. McMurray \cite{mcmurray2013poultry} examined the unique constraints of poultry processing and listed enormous variability among carcasses, breast meat that was easily deformable and easily bruised. Although this gripper improved throughput, issues still remained due to size differences, slippery, and irregularly positioned products. 


These challenges underscore the need for more adaptive grippers that could handle deformable poultry products with greater flexibility and reliability. The latest advances in automating meat processing have led to technologies such as the Meat Factory Cell (MFC), in which Mason et al. \cite{mason2021meatfactory} employed AI-driven robotic methods to grasp, cut up, and manipulate carcasses with increased efficiency. Such methods, however, are primarily rule-based automation without feedback interactions between the system and the deformable products. Zhou et al. \cite{zhou2007shoulder} illustrated a force-controlled robot deboning system with resizing adjustment mechanisms, but their method relied on hard-coded paths and was not able to accommodate product variance. As an alternative, Ahlin \cite{ahlin2022workbench} suggested the Robotic Workbench—a general-purpose approach to poultry automation that attempts to deal with product variations using sensor-based controls. While this system improves responsiveness compared to traditional methods, it still employs predefined, rule-based operations and is not capable of dynamically varying its manipulation strategies when faced with the large variations in poultry products.


Beyond the meat processing industry, AI-powered robots have significantly advanced the agricultural industry. For example, You et al. \cite{you2021pruning} combined vision-based control with admittance feedback to achieve precise orchard pruning, while Yaguchi et al. \cite{yaguchi2016tomato} developed a stereo camera and rotational plucking system for tomato harvesting. Silwal et al. \cite{silwal2021bumblebee} introduced a fully autonomous grapevine-pruning platform, and Xiong et al. \cite{xiong2019strawberry} tackled strawberry harvesting with active obstacle separation and dual-arm pick-and-place. Each of these solutions made progress in perception or motion optimization, but they typically rely on rule-based or classical control loops. Their successes demonstrate the potential of robotics in agriculture but also highlight limitations in rigid control systems handling unforeseen scenarios—a key motivation for deeper learning-based methods that can adapt on the fly.


To fill the demand for dynamic handling the deformable objects, imitation learning, a learning-based robotic control strategy, initially trains the robot by a human operator's demonstration of the task. Imitation learning, also known as Learning from Demonstration (LfD), has its roots in foundational work like Pomerleau’s ALVINN \cite{pomerleau1988alvinn}, which demonstrated that a neural network trained on human driving data could outperform rigid, rule-based control approaches. For an overarching view of imitation learning’s evolution and methods, see \cite{hussein2018survey}. Subsequent research by Atkeson and Schaal \cite{atkeson1997demonstration} showed that by combining model-based planning with LfD, trained robots can accomplish complex manipulation tasks. Modern extensions, such as the Diffusion Policy \cite{chi2024diffusion}, improved upon traditional policy representations by viewing robot control as a conditional denoising process, enhancing stability, accommodating multimodal actions, and refining performance in high-dimensional tasks. Moreover, widely adopted vision models \cite{ronneberger2015unet} have proven to be invaluable for robust perception and segmentation, underscoring how advanced deep-learning methods can improve the image data understanding and bolster the adaptability of imitation-learning frameworks in robotics.


In recent years, imitation learning has shown early-stage success by being able to handle easily some deformable food products. Kim et al. \cite{kim2024pepper} trained a visuomotor policy for outdoor pepper harvesting, demonstrating the feasibility of imitation learning under unstructured field conditions. Liu et al. \cite{liu2024forcemimic} integrated a force-aware data collection control system into an imitation learning framework (ForceMimic). By combining these two systems, there was a significant improvement in the success rates in contact-rich tasks like zucchini peeling. Similar sensor-rich and data-driven approaches have been used in dual-arm banana peeling \cite{kim2024dualaction}, one-shot mushroom harvesting \cite{porichis2024imitation}, and apple picking \cite{vandeven2024apple}. These studies demonstrate how imitation learning can compensate for unstructured environments, complex object morphologies, and delicate or deformable products. Future work could also explore full-body motion imitation, as in DexMV \cite{xia2022dexmv}, which has been shown to be able to capture more nuanced manipulations. A comprehensive survey \cite{mahmoudi2024survey} further underscores the advantages and challenges of bringing imitation learning into real-world agricultural settings, noting that data quality, environmental variability, and robust policy designs still remain unresolved issues. In parallel, Hovakimyan \cite{hovakimyan2022crop} explored the combined use of reinforcement learning and imitation learning for optimizing crop management, highlighting the potential of hybrid learning-based strategies in agriculture. Although recent imitation learning studies have succeeded on relatively rigid crops, extending these techniques to poultry processing exposes two domain-specific challenges:




\begin{enumerate}
    \item \textbf{Deformability and slipperiness:} The skin shears and stretches under load, while surface moisture prevents suction seals.
    \item \textbf{Anatomical variability:} Leg spacing, length, and orientation differ across birds, demanding centimetre-scale compliance.
\end{enumerate}

Building on these limitations, this paper presents a novel approach named ChicGrasp, which is a dual-jaw pneumatic gripper with an imitation learning-based approach tailored to the unique demands of poultry processing, particularly handling easily deformable and variable broiler carcasses. By boot-strapping from expert demonstrations, imitation learning (IL) avoids the exhaustive edge-case enumeration of rule-based programs and the sparse-reward, safety-reset difficulties of reinforcement learning.

Key contribution of this work is summarized as follows: (i) an independent 2-DoF dual-jaw gripper that clamps both legs without suction or bruising and (ii) a conditional diffusion policy that jointly predicts 5-DoF robot motion and binary jaw states from RGB + proprioception. On real broiler carcasses, ChicGrasp achieved 41 / 101 successful pick-and-rehang trials (40.6\%), whereas IBC and LSTM-GMM baselines achieved 0\%. All CAD files, code, 50 tele-operation trajectories, and evaluation logs will be released to accelerate ag-robotics research. By overcoming the limitations of static, rule-based systems, our solution offers a scalable, adaptable alternative that can reduce manual labour, raise throughput, and improve handling consistency in poultry processing. The remainder of this paper is organized as follows: Section II details the gripper design and diffusion-policy controller, Section III presents real-world robotic demonstrations, Section IV discusses the findings, and Section V concludes with directions for future work.

\section{Method}

\begin{figure}[!h]
    \centering
        \includegraphics[width=3.5in]{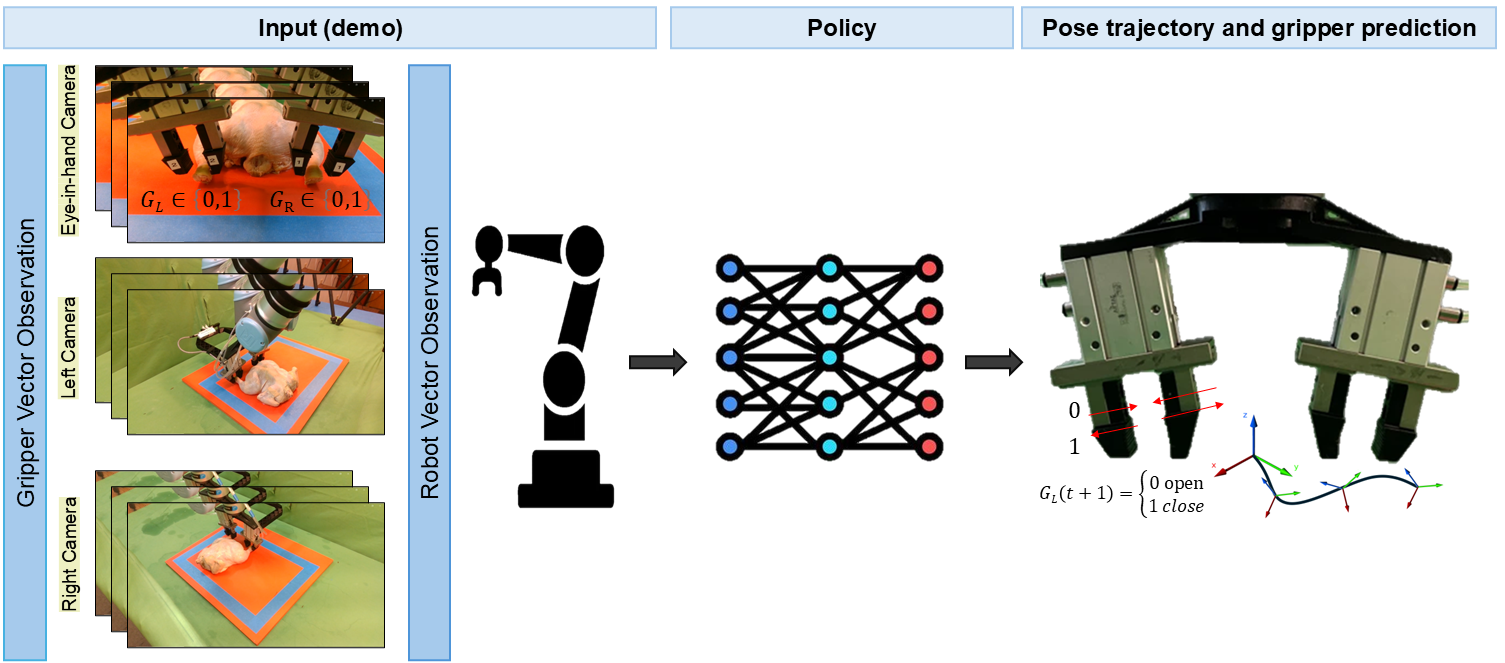}
        \caption{Learning a grasping policy from demonstration: The network inputs stacked image frames, robot state vectors, and gripper states, and outputs both the end‐effector pose trajectory and gripper control commands.}
    \label{Fig3_architecture}
\end{figure}

This section describes the hardware design, data collection pipeline, learning algorithms, and runtime control policy that collectively enable the \emph{ChicGrasp} system to pick and rehang chicken legs using imitation learning. A high-level summary of the flow of information is given in Figure \ref{Fig3_architecture}; specific hardware and learning components are described in the following subsections, from sensing and actuation to policy execution.

\subsection{Customized End-Effector Design}
\begin{figure}[!h]
    \centering
        \includegraphics[width=3in]{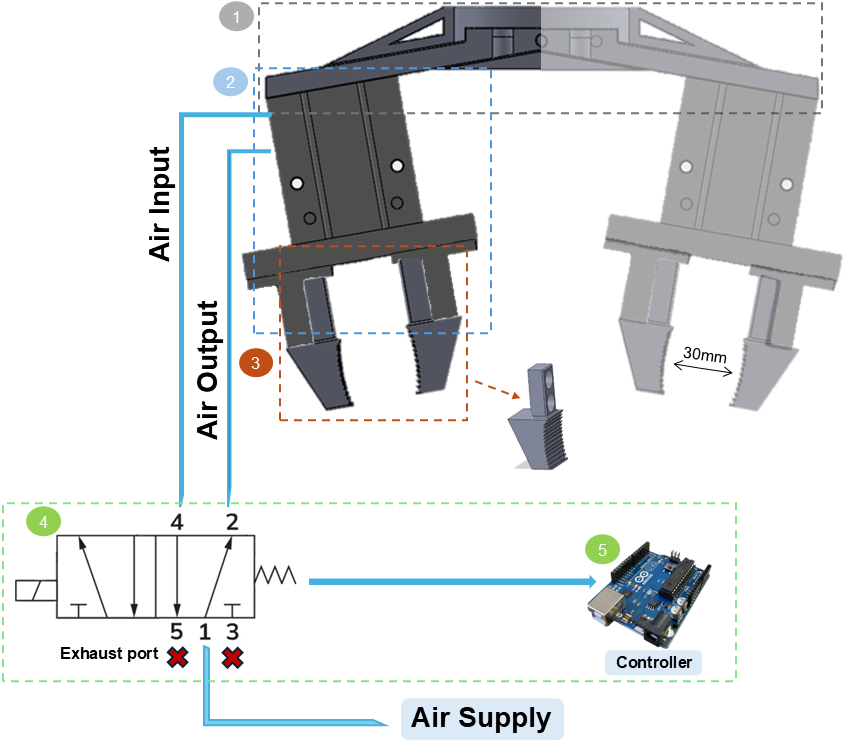}
        \caption{Schematic of the 2DoF customized gripper system for chicken carcass grasping. The system consists of: (1) a 3D-printed interface for robot attachment, (2) a pneumatic actuator for jaw control, (3) 3D-printed serrated jaw surface (30° chevron ridges), (4) a pneumatic valve regulating air input and output, and (5) a control section incorporating a microcontroller for system operation. Air supply is routed through the valve to actuate the pneumatic gripper, enabling controlled gripping and releasing motions.}
    \label{Fig4_gripper}
\end{figure}

The primary hardware innovation is a custom-built, independently actuated, pneumatically driven dual-jaw gripper, weighing approximately 4 kg. The gripper is designed to conform to the irregular and slippery geometry of chicken legs without causing tissue damage. Both jaws translate linearly under pneumatic actuation and terminate in additively-manufactured chevron ridges (30° pitch, 2 mm depth) that convert part of the normal clamping force into tangential resistance, thereby increasing friction on wet, deformable skin while avoiding tissue piercing. The actuation is powered by double-acting parallel air grippers (Airtac HFZ series, China), controlled via a 5/2-way solenoid valve (Tailonz 4V210-08, USA). Valve states are toggled by an \mbox{Arduino Uno R4}, which receives high-level commands from the host PC through a USB Type-C connection. A schematic of the full pneumatic gripper system, including actuator, control, and air flow components, is shown in Figure \ref{Fig4_gripper}. To enable effective data collection for training the diffusion policy model, an eye-in-hand camera is mounted above the gripper using a 3D-printed holder. This holder is designed to ensure that both jaws remain in the camera's field of view during grasping, allowing for consistent visual capture of jaw behavior across demonstrations. The gripper does not close both jaws simultaneously; each jaw actuates independently. During operation, the right jaw closes as soon as it aligns with the right leg, and the left jaw follows once it reaches its position. This staggered sequence accommodates asymmetric leg spacing, prevents one jaw from displacing the carcass before the other engages, and increases robustness to positional variation, resulting in a more stable grasp.

To support learning from demonstration, it is important to capture the state of the gripper over time. Since each jaw operates independently in a binary open/closed manner, we encode their state using binary variables, gripper left, gripper right \( g_L,\, g_R \in \{0, 1\} \), where \( g_{L} = 1 \) (or \( g_{R} = 1 \)) signifies that the left (or right) jaw is closed, and \( 0 \) indicates it is open. Over a finite planning horizon of \( T \) control steps, the temporal evolution of the jaws is denoted as \( g_{L}(t),\, g_{R}(t) \in \{0, 1\} \) for \( t = 1, \dots, T \). These states are stored alongside proprioception and vision so that the learning algorithm can benefit from causal dependencies between jaw actuation and future observations.

\subsection{Action Space and Policy Formulation}
\label{sec:action}


The manipulator–gripper pair is controlled using a five-dimensional action vector defined as
\begin{equation}
\mathbf{a}_{t} =
[x_{t},\, y_{t},\, z_{t},\, g_{L,t},\, g_{R,t}]^{\top}, 
\label{eq:action_vector}
\end{equation}
where \( (x_{t},\, y_{t},\, z_{t}) \) denote the end-effector position in the robot base frame $t$, and \( g_{L,t} \), \( g_{R,t} \) are the binary jaw variables defined earlier.

A conditionally denoising diffusion model \cite{chi2024diffusion} with parameters $\theta$ is trained to map the current observation $\mathbf{s}_{t}$, which consists of stacked RGB frames, joint positions, and jaw states to a distribution over Equation \eqref{eq:action_vector}. Recent goal-conditioned approaches leverage score-based diffusion for flexible policy generation \cite{chen2023goal}. The output is then denoised and sampled, and the resulting continuous jaw logits are binarized through the following equation:


\begin{equation}
    \tilde{g}_{\cdot, t} = 
    \begin{cases}
    1, & \text{if } \sigma(\hat{g}_{\cdot, t}) > 0.5 \\[2pt]
    0, & \text{otherwise}
    \end{cases}
    \label{eq:threshold}
\end{equation}
where $\sigma(\cdot)$ denotes the logistic sigmoid function.

\begin{figure}[!h]
    \centering
        \includegraphics[width=2in]{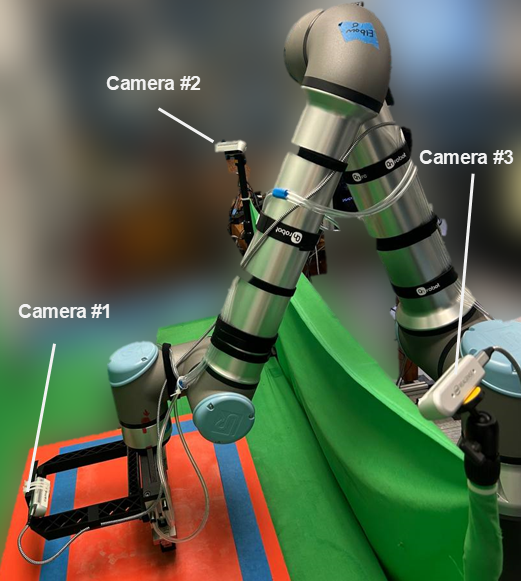}
        \caption{Experimental setup for multi-view data collection. Three RGB cameras are positioned to capture different perspectives of the manipulation space: Camera 1 is mounted directly on the gripper (eye-in-hand configuration) to provide a dynamic front view of the target object, and Camera 2 and Camera 3 are placed on the right and left side of the table, respectively, for a side-overview perspective.}
    \label{camera setup}
\end{figure}

To facilitate robust visual perception, the setup employs two RGB cameras positioned on the left and right sides of the workspace (Camera 2 and Camera 3 in Figure~\ref{camera setup}). This dual-view setup mitigates occlusions and captures complementary perspectives of the chicken carcass, especially when body parts like legs are not fully visible from a single angle. The image sequences from these cameras are stacked and included in the observation vector $\mathbf{s}_t$, providing the policy with temporally and spatially rich input for accurate grasp planning.


The network is trained using an $\ell{2}$ imitation loss through Equation \eqref{eq:loss}:

\begin{equation}
\mathcal{L}(\theta) = \mathbb{E}_{(\mathbf{s}, \mathbf{a}) \sim \mathcal{D}} \left\| \mathbf{a} - \hat{\mathbf{a}}_{\theta}(\mathbf{s}) \right\|^2,
\label{eq:loss}
\end{equation}

where $(\mathbf{s},\mathbf{a})$ pairs are drawn from the demonstration dataset $\mathcal{D}$.

We evaluate three policies that generate actions in the same space defined by Equation~\eqref{eq:action_vector}: (i) diffusion policy \cite{chi2024diffusion}, which uses a denoising diffusion model to generate actions conditioned on visual and proprioceptive observations; (ii) an LSTM-GMM \cite{mandlekar2021whatmatters} baseline that autoregressively predicts a Gaussian mixture over $\mathbf{a}_{t}$; and (iii) implicit behavioral cloning (IBC) \cite{Florence2022IBC} with an energy-based model. All three policies share the perception backbone introduced in the Diffusion-Policy paper \cite{chi2024diffusion}: a ResNet-18 encoder \cite{he2016resnet} (trained from scratch) with Spatial Softmax \cite{mandlekar2021whatmatters} and GroupNorm \cite{wu2018group} that processes stacked RGB frames end-to-end.



\subsection{Demonstration Collection}
\label{sec:data}

The demonstrations were collected using the teleoperation method. The UR10e was guided by a master operator using a 3Dconnexion\textsuperscript{\textregistered} SpaceMouse to lift and gently grasp chicken legs that were set on a red plastic table. The synchronous data streams accompanying these were: 1) \textbf{Vision:} three Depth Cameras D435 capturing $640\times480$ RGB at 30 Hz; 2) \textbf{Proprioception:} joint positions and velocities at 100 Hz from the robot controller; and 3) \textbf{Gripper:} binary jaw positions at 100~Hz sampled by the Arduino.


\begin{figure*}[t]
    \centering
    \includegraphics[width=\textwidth]{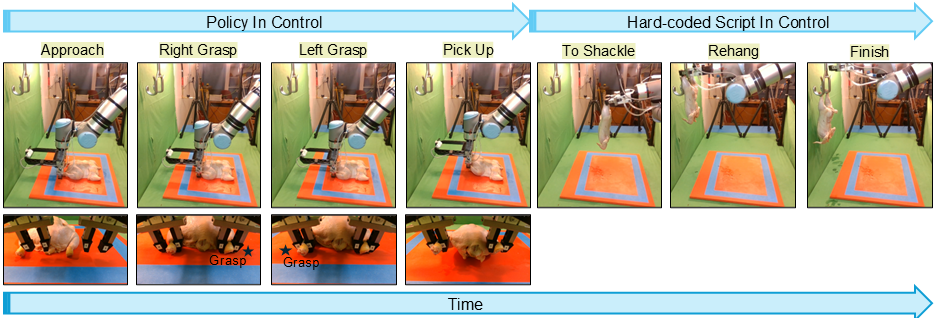}
    \caption{Sequential grasping and pickup actions of the robotic system over time.}
    \label{Fig4_timesteps_pick_rehang}
\end{figure*}

\begin{figure}[!h]
    \centering
        \includegraphics[width=3.5in]{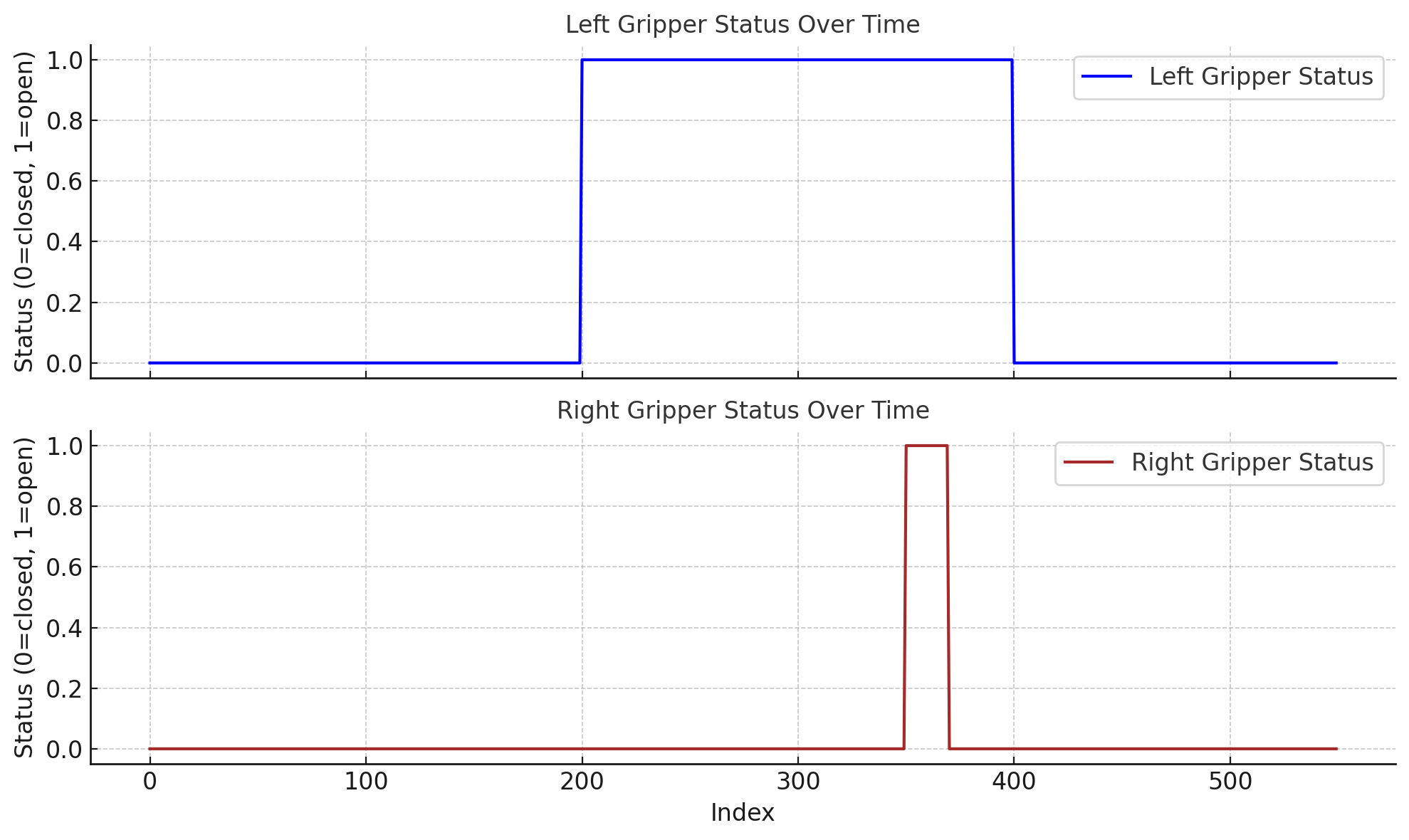}
        \caption{Temporal Variation of Left and Right Gripper States: The plots depict the binary state transitions (0 = Closed, 1 = Open) of the left and right jaws of the gripper over time. The data highlights the activation patterns and synchronization of the gripper during operation.}
    \label{Fig5_gripper_low_data}
\end{figure}

\begin{figure*}[!h]
    \centering
    \begin{minipage}{\linewidth}
        \centering
        \includegraphics[width=5in]{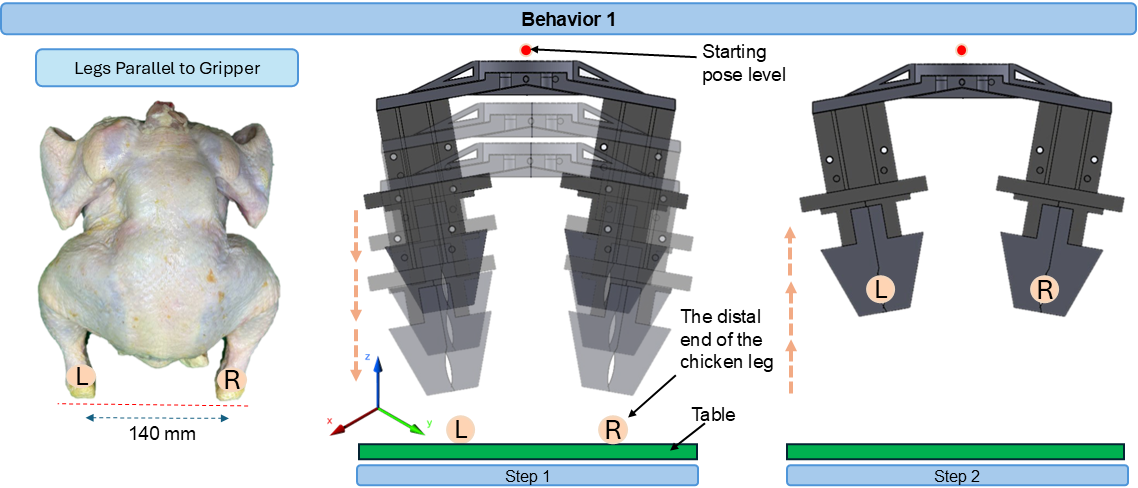}
        \caption{\textbf{Left}: Top-down perspective of a chicken carcass facing breast down, whose left (L) and right (R) legs are roughly 140 mm apart and parallel to the opening of the gripper.
        \textbf{Center \& right}: Two-step sequence of grasp being carried out by dual-jaw end-effector. Throughout Step 1, the gripper goes downwards from the original pose (red dot) until every jaw encompasses the distal part of the leg at table height. Throughout Step 2, the jaws close simultaneously, and the carcass is vertically picked up along the dashed route, beginning the pick-and-rehang operation. Coordinate axes denote the robot frame; the green bar is the table surface.}
    \label{Fig8_demo_steps.png}
    \end{minipage} 
\end{figure*}

A total of $N=50$ trajectories (25-40 s each) were recorded that span diverse carcass poses and lighting conditions. Figure \ref{Fig4_timesteps_pick_rehang} illustrates sequential grasping and pickup actions for the robotic system, where the picking stage is using imitation learning and the rehang stage is using a hard-coded script, while the temporal jaw profiles are summarized in Figure \ref{Fig5_gripper_low_data}. Figure \ref{Fig8_demo_steps.png} illustrates a top-down grasping scenario, showing how the gripper jaws align with the chicken legs prior to closure.

\subsection{Policy Training}
\label{sec:training}
Training details are summarized in Table \ref{tab:table3}.  We used different batch sizes because each method has distinct memory/performance trade-offs: Diffusion Policy consumes more VRAM per sample due to its temporal UNet, limiting feasible batch size to 32 on an 80 GB A100; IBC benefits from large negative-sampling batches and therefore ran with 128 on a 32 GB V100; LSTM-GMM sits in-between at 64. All models were trained for 600 epochs.


\subsection{Runtime Control Loop}
\label{sec:runtime}

At each inference step the policy observes the state vector $\mathbf{s}_t\!=\ [\mathbf{o}_t^{\text{img}},\, \mathbf{q}_t^{\text{joint}},\, g_{L,t},g_{R,t}]$, where $\mathbf{o}_t^{\text{img}}$ are stacked RGB frames, $\mathbf{q}_t^{\text{joint}}$ are the 6-DoF joint positions, and $g_{L,t},g_{R,t}\!\in\!\{0,1\}$ are the binary jaw flags. It then samples an action $\mathbf{a}_t\!=\ [x_t,\,y_t,\,z_t,\,g_{L,t}^\star,\,g_{R,t}^\star]$ and applies the threshold rule in Equation~\eqref{eq:threshold}. The first three scalars $(x_t,y_t,z_t)$ are streamed to the robot velocity controller at 250\,Hz, while the two bits $(g_{L,t}^\star,g_{R,t}^\star)$ are forwarded to an Arduino Uno, which toggles a 5/2-way solenoid valve to open/close each jaw. Once the policy predicts a stable grasp, defined as $g_{L,t}^\star\!=\!g_{R,t}^\star\!=\!1$ for three consecutive frames, the controller switches to a pre-computed, seven-waypoint trajectory that lifts the carcass and moves it to the overhead shackle.  This hybrid strategy (learned grasp \(\rightarrow\) scripted rehang) guarantees collision-free motion after the bird has been lifted and is no longer in contact with the table. In future work, we plan to replace the scripted segment with imitation learning as well, using a better teleoperation method such as a customized motion-capture gripper. 





For notation, we define $g_{L,R,t} \triangleq (g_{L,t}, g_{R,t})$ as the pair of left and right jaw states at time $t$.
Mathematically, the closed-loop dynamics are

\begin{equation}
\mathbf{s}_{t+1}=f(\mathbf{s}_t,\mathbf{a}_t),\label{eq:dynamics}
\end{equation}

where $f(\cdot)$ concatenates the continuous forward kinematics of the robot with the discrete valve-logic update $g_{L,R,t+1}=g_{L,R,t}^\star$. 
Thus, $f$ is a hybrid transition model that jointly evolves robot pose and gripper state.





\section{Results}

We experimented with all three models. The task was to pick and rehang three different chicken carcass exemplars of different sizes and weights by a UR10e robot with a specially created dual-jaw pneumatic gripper with 2 degrees of freedom. There were 50 pick-and-rehang trials per method across the exemplars. Table \ref{tab:diffusion_ibc_results} summarizes the outcomes: 1) diffusion policy posted a total success rate of 40.6\% (41 successful lifts from 101 tries). Per-exemplar rates ranged from 25.3\% (14/55) to 64.5\% (20/31). Each successful pick-and-rehang cycle, from the first policy action to the final release at the shackle, was around 38 seconds. Speed limitation: this is an extremely important limitation in the current setup that needs to be addressed. The autonomous system is not yet comparable to a human worker, who can manually hang birds on a shackle line that operates at 140 birds per minute (bpm), approximately one bird every 0.43 seconds; 2) The IBC failed, achieving a success rate of 0\%. It took around 60 seconds per attempt on average, primarily because the policy failed to commit to a specific trajectory and often timed out. Its motion patterns were erratic and unstructured, making it unable to locate the chicken reliably; and 3) LSTM-GMM also failed to complete any successful grasp-and-rehang cycles. However, it performed better than IBC in terms of trajectory stability. The robot, under LSTM-GMM, was able to smoothly navigate toward the chicken legs but consistently failed to actuate the gripper jaws after reaching the correct position. In contrast, IBC's movements were unpredictable and aggressive, often failing to even find the target legs.

\begin{figure}[!h]
    \centering
        \includegraphics[width=3.5
        in]{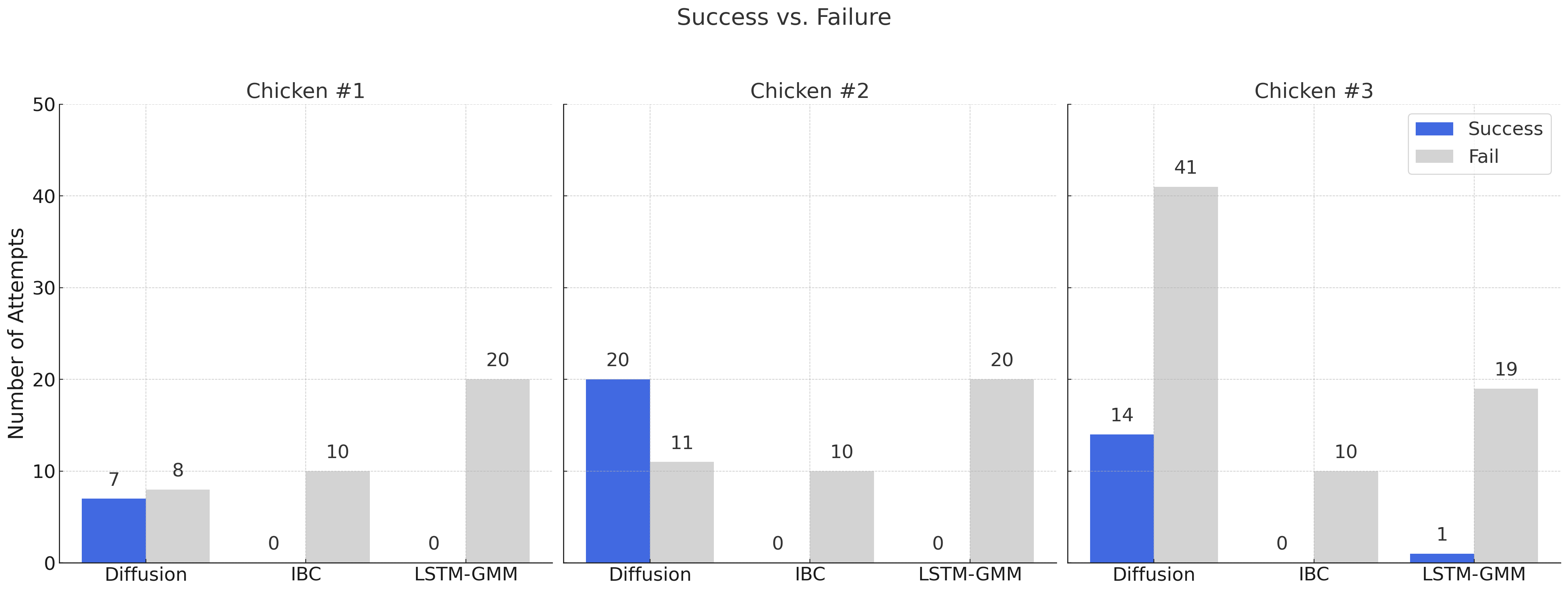}
        \caption{Success vs. failure comparison for diffusion policy, implicit behavioral cloning (IBC), and LSTM-GMM.}
    \label{fig_9}
\end{figure}

Success was defined by three conditions: (i) clamping the two chicken legs between the jaws, (ii) lifting the carcass at least 50 mm without slipping, and (iii) rehanging the carcass at the overhead shackle by a fixed waypoint path. Figure \ref{fig_9} illustrates success over failure frequencies for both models in the three exemplars. The chicken legs are nearly perpendicular to the gripper approach direction, and the legs are at the same level, facilitating a simple top-down grasp.

\begin{table*}[h]
    \caption{Evaluation results for Diffusion Policy, IBC, and LSTM-GMM across three chicken exemplars.}
    \label{tab:diffusion_ibc_results}
        \centering
        \begin{tabular}{@{}lllllll@{}}
        \toprule
        \textbf{Methods} & \textbf{Idx} & \textbf{Success Rate} & \textbf{Successful} & \textbf{Failed} & \textbf{Total} & \textbf{Exec. Time} \\
        \midrule
        \multirow{4}{*}{Diffusion Policy} 
            & Chi\_1 & 46.67\%  & 7  & 8  & 15  & 28 Sec \\
            & Chi\_2 & 64.52\%  & 20 & 11 & 31  & 28 Sec \\
            & Chi\_3 & 25.25\%  & 14 & 41 & 55  & 28 Sec \\
            & \textbf{Total} & \textbf{40.59\%} & \textbf{41} & \textbf{60} & \textbf{101} & \textbf{} \\
        \midrule
        \multirow{4}{*}{IBC} 
            & Chi\_1 & 0\%      & 0  & 10 & 10  & 60 Sec \\
            & Chi\_2 & 0\%      & 0  & 10 & 10  & 60 Sec \\
            & Chi\_3 & 0\%      & 0  & 10 & 10  & 60 Sec \\
            & \textbf{Total} & \textbf{0\%} & \textbf{0} & \textbf{30} & \textbf{30} & \textbf{} \\
        \midrule
        \multirow{4}{*}{LSTM-GMM} 
            & Chi\_1 & 0\%      & 0  & 20 & 20  & 60 Sec \\
            & Chi\_2 & 0\%      & 0  & 20 & 20  & 60 Sec \\
            & Chi\_3 & 5\%      & 1  & 19 & 20  & 60 Sec \\
            & \textbf{Total} & \textbf{0\%} & \textbf{1} & \textbf{29} & \textbf{60} & \textbf{} \\
        \bottomrule
        \end{tabular}
\end{table*}


To evaluate generalization across natural variability, we conducted experiments on three different broiler carcasses, referred to as Chicken 1, Chicken 2, and Chicken 3. These exemplars differed in size, leg spacing, and surface condition, reflecting typical variations found in poultry processing lines. In addition, some carcasses were extremely fresh, showing normal skin coloration, while others were tested after some aging, which caused noticeable changes in skin texture and color.

Each panel reports the grasp results for a different carcass exemplar (Chicken 1–3). Within each panel, the blue bars show the number of successful pick‑and‑lift attempts and the grey bars the failed attempts for (i) the proposed Diffusion Policy controller, (ii) the implicit behavioral cloning (IBC) baseline, and (iii) the LSTM-GMM baseline. The diffusion policy achieved 7/15, 20/31 and 14/55 successes in Chickens 1 to 3, respectively. IBC registered 0/10 successes for all exemplars. LSTM-GMM also failed to consistently grasp and rehang, posting 0/20, 0/20, and 1/20 successes for Chickens 1 to 3, respectively. Table~\ref{tab:diffusion_ibc_results} provides a summary of task configurations and execution modes.


\begin{table}[!h]
    \centering
    \caption{Overview of model training statistics.}
    \label{tab:table3}
    \begin{tabular}{lccccc}
        \toprule
        \textbf{Model} & \textbf{Epochs} & \textbf{Batch Size} & \textbf{GPU} \\ 
        \midrule
        Diffusion & 600 & 32 & A100 (80 GB) \\
        IBC       & 600 & 128 & V100 (32 GB) \\
        LSTM-GMM       & 600 & 64 & V100 (32 GB) \\
        \bottomrule
    \end{tabular}
\end{table}


\section{Discussion}


The performance gap between diffusion policy and IBC demonstrates the influence of strong action distributions on imitation learning, especially in challenging cases involving deformable, slippery objects such as raw poultry carcasses. Representing actions as conditional distributions over end-effector poses and jaw states, the diffusion framework achieved better results than the energy-based IBC approach. The LSTM-GMM baseline also struggled, achieving almost no successful grasps and highlighting the limitations of traditional sequence modeling techniques in contact-rich deformable object manipulation. Yet, diffusion policy's success ratio, though higher than that of IBC and LSTM-GMM, is at best modest (40.6\%).

Several limitations contributed to the observed modest success rate. 
Addressing these challenges points to several opportunities for future improvement. 1) 50 teleoperated demonstrations were sampled. The relatively small data set might not capture the entire spectrum of carcass orientations and surface frictions. A more diverse sample of carcass states (e.g., different sizes, orientations, and variation in leg position) can enhance policy robustness; 2) While the custom dual-jaw gripper is able to reliably enclose both legs, the learned policy sometimes drives the end-effector to a point slightly below the ideal grasp location. Consequently, the legs are picked from a suboptimal vertical offset, complicating stable transfer and alignment during the following rehang stage; 3) Additional real-time feedback (e.g., tactile sensing or force feedback) about slippage or partial grasps would greatly enhance corrective behavior; 4) The overhead shackle rehang script functioned consistently after the carcass was picked. Most failures occurred at the pick stage: if the grasp was unstable or incomplete, it resulted in an instantaneous drop of the carcass. A more subtle strategy that mingles learning-based grasping with partially learned rehang movements may be able to increase the success even further (e.g., using motion capture teleportation); and 5) The average cycle time of around 38 seconds for Diffusion Policy primarily reflects conservative end-effector velocities and the overhead scripted rehang sequence. While this duration is acceptable for a proof-of-concept demonstration, it is far from meeting the pace required for industrial applications. For instance, commercial poultry processing lines operate at approximately 140 birds per minute, equivalent to one bird every about 0.43 seconds. In such settings, human workers typically complete the rehang task in under 2 seconds. This speed gap represents a critical bottleneck in the current system. Future iterations should investigate motion optimization strategies, such as faster policy rollouts, higher joint velocities, and model-predictive or closed-loop control, to reduce latency and bring autonomous performance closer to industrial throughput requirements.

Despite these limitations, the successful trials demonstrate the advantage of learned flexible policies in treating deformable, irregular bio-products. The diffusion-based method better encoded spatiotemporal relationships among vision, proprioception, and binary jaw states to enable better generalization over test conditions than both the IBC and LSTM-GMM baselines.


\section{Conclusion}

The primary objective of this study was to develop and demonstrate a proof-of-concept system for automated chicken carcass manipulation using imitation learning. We introduced a dual-jaw pneumatic gripper tailored for delicate, deformable objects and integrated it with a diffusion policy trained from multi-view teleoperation data. Our key findings include: 1) \textbf{Feasibility:} A learned Diffusion Policy successfully grasped and rehung irregular poultry carcasses at a non-trivial rate (40.6\% overall), outperforming an IBC baseline that entirely failed; 2) \textbf{Policy-Driven Grasping:} By leveraging a conditional diffusion model, the robot captured subtle cues from RGB images and proprioceptive feedback, enabling the effective operation of both gripper jaws; and 3) \textbf{Hybrid Approach:} Coupling a learned pick policy with a scripted rehang motion served as a stepping-stone strategy, ensuring consistent shackling once the carcass was securely lifted. However, we plan to capture rehang demonstration data in future work, allowing the entire pick-and-rehang sequence to be learned end-to-end.


Looking forward, we identify several avenues for improvement: increasing training data diversity to enhance generalization, fine-tuning the gripper or pickup strategies for optimal leg enclosure, integrating closed-loop sensors for real-time corrective maneuvers, and optimizing the motion plan to meet tighter industrial cycle times. Overall, these results demonstrate that integrating task-specific hardware with advanced policy learning can significantly reduce manual labor in challenging handling tasks in the poultry processing industry.  
Future research will concentrate on expanding the system to other delicate food products outside of poultry, adding more adaptive grasping behaviors, and scaling it to handle faster line speeds.  
By doing this, we hope to close the gap between automation proof-of-concept and industrial implementation in actual food processing settings.

\section{Acknowledgement}
This work was supported by awards No. 2023-67021-39072, 2023-67022-39074, and 2023-67022-39075 from the U.S. Department of Agriculture (USDA)’s National Institute of Food and Agriculture (NIFA) in collaboration with the National Science Foundation (NSF) through the National Robotics Initiative (NRI) 3.0.

\bibliographystyle{IEEEtran}
\bibliography{references}
\end{document}